\pdfoutput=1

\documentclass[11pt]{article}

\usepackage[]{ACL2023}

\usepackage{times}
\usepackage{latexsym}
\usepackage{ragged2e}
\usepackage{multirow}
\usepackage{graphicx}
\usepackage{caption}
\usepackage{subcaption}
\usepackage[english]{babel}
\babelprovide[import]{thai}

\usepackage{graphicx}
\usepackage{dblfloatfix}
\usepackage{subcaption}
\usepackage{tablefootnote}
\usepackage{multicol}
\usepackage{multirow}
\usepackage{adjustbox}
\usepackage{comment}
\usepackage{placeins}
\usepackage{caption}
\usepackage{subcaption}
\usepackage{mathtools}
\usepackage{amsmath}
\usepackage{algorithm}
\usepackage{soul}
\usepackage{algpseudocode}
\algblock{Input}{EndInput}
\algnotext{EndInput}
\algblock{Parameters}{EndParameter}
\algnotext{EndParameter}

\usepackage[T1]{fontenc}

\usepackage[utf8]{inputenc}

\newcommand{\ankan}[1]{\textcolor{red}{[\bf\small ankan?: #1]}}

\usepackage[T1]{fontenc}

\usepackage[utf8]{inputenc}

\usepackage{microtype}

\usepackage{inconsolata}

\newcommand{\dom}{Dominant\xspace}
\newcommand{\nondom}{Non-Dominant\xspace}
\newcommand{\mlmcc}{\textbf{\emph{MLMCC}}}
\newcommand{\mlmcid}{\textbf{\emph{MLMCID}}}

\usepackage[utf8]{inputenc}

\title{A Pointer Network-based Approach for Joint Extraction and Detection of Multi-Label Multi-Class Intents}

\author{$^1$Ankan Mullick \qquad $^1$Sombit Bose \qquad $^1$Abhilash Nandy \qquad \\{\bf $^2$Gajula Sai Chaitanya \and $^1$Pawan Goyal} \\ \texttt{\{ankanm, sbcs.sombit.24, nandyabhilash\}@kgpian.iitkgp.ac.in} \\ \texttt{gsaichai@qti.qualcomm.com} \qquad \texttt{pawang@cse.iitkgp.ac.in}\\ $^1$Computer Science and Engineering Department, IIT Kharagpur, India \qquad $^2$Qualcomm, India}

\begin{document}

\maketitle

\begin{abstract}

In task-oriented dialogue systems, intent detection is crucial for interpreting user queries and providing appropriate responses. Existing research primarily addresses simple queries with a single intent, lacking effective systems for handling complex queries with multiple intents and extracting different intent spans. Additionally, there is a notable absence of multilingual, multi-intent datasets. This study addresses three critical tasks: extracting multiple intent spans from queries, detecting multiple intents, and developing a multilingual multi-label intent dataset. We introduce a novel multi-label multi-class intent detection dataset (\textbf{MLMCID-dataset}) curated from existing benchmark datasets. We also propose a pointer network-based architecture (\textbf{MLMCID}) to extract intent spans and detect multiple intents with coarse and fine-grained labels in the form of sextuplets. Comprehensive analysis demonstrates the superiority of our pointer network based system over baseline approaches in terms of accuracy and F1-score across various datasets.

\end{abstract}

\begin{figure*}[!th]
    \centering
    \vspace{-5mm}
    \includegraphics[width=\textwidth]{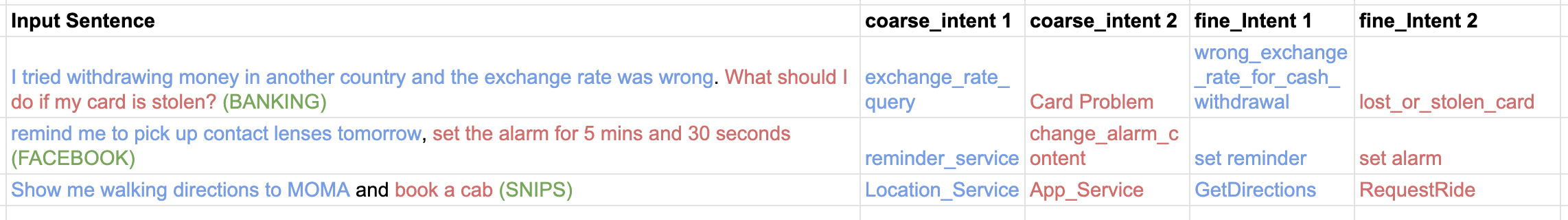}
    \vspace{-4mm}
    \caption{Examples of multi-label multi intent datasets (SNIPS, Facebook and BANKING) }
    \label{fig:multi-intent-example}
    \vspace{-2mm}
\end{figure*}

\section{Introduction}



Task-oriented dialogue systems have become a major field of study in recent years, significantly advancing the capabilities of Natural Language Understanding (NLU). These systems execute command-based tasks, demonstrating versatility in handling diverse user queries through a set of predefined skills, known as intents. Users interact with dialogue systems to fulfill their needs, and intent detection plays a pivotal role in comprehending user queries and generating appropriate responses in task-oriented conversations, thereby maintaining user engagement.
The task of intent detection involves identifying the \textit{intent(s)} within a given statement or query, which represents the underlying meaning conveyed by the user. For example, the query ``How is the weather today?" would be associated with the $GetWeather$ intent. Dialogue systems rely on detecting these intents to understand user queries and provide suitable answers.


However, in real-world conversation, a query or a statement often contain multiple different intents. For instance, as shown in Fig. \ref{fig:multi-intent-example}, for the query (from Facebook English dataset): ``remind me to pick up contact lenses tomorrow, set the alarm for 5 mins and 30 seconds", contains two distinct intent categories with following spans: `remind me to pick up contact lenses tomorrow' (`set reminder' intent) and `set the alarm for 5 mins and 30 seconds' (`set alarm' intent). Both of these are fine intent categories. Multiple similar fine intents can be merged to create one coarse intent as explained in Table \ref{tab:course-fine-data-fb}. Thus, the above query contains `reminder\_service' and `change\_alarm\_content' coarse intents as shown in Fig. \ref{fig:multi-intent-example}. In case of multiple intents in a sentence, one intent which is dominant and most important in that sentence can be termed as `Primary' intent while the other intents can be considered `Non-Primary'. For example, in the query (From Mix-SNIPS dataset) ``How is the weather today? It would be lovely to go for a movie" is a combination of two simple sentences `How is the weather today?' and `It would be lovely to go for a movie', whose intents are $GetWeather$ and $BookMovieTicket$ respectively. Out of the two possible intents, $BookMovieTicket$ is primary (primary and main focus of the sentence) and $GetWeather$ becomes non-primary. It would require an intent span extraction algorithm to extract multiple intent spans and a multi-label, multi-class classifier to detect different fine and coarse intents.

Over the past few years, researchers concentrate on intent identification across different domains. Flexible and adaptive intent class detection models have been developed for dynamic and evolving real-world applications. \cite{liao2023novel,kuzborskij2013n, scheirer2012toward, degirmenci2022efficient} focus on streaming data to identify evolving new classes using incremental learning. SENNE \citet{cai2019nearest}, IFSTC \cite{xia2021incremental}, SENC-MaS \cite{mu2017streaming}, SENCForest \cite{mu2017classification}, ECSMiner \citep{masud2010classification} aim at SENC (streaming emerging new class) problem on intents on streams. \cite{sun2016online} work on emergence and disappearance of intents. 
\cite{wang2020active} uses high dimensional data for streaming classification. \cite{mullick2022framework} identifies multiple novel intents using a clustering framework. \cite{na2018dilof,zhan2021out,larson2019evaluation,yan2020unknown, zhou2022knn, firdaus2023multitask} detect new intents in the form of outlier detection. Unlike the previous single-intent detection models, which can easily utilize the utterance's sole intent to guide slot prediction, multi-intent SLU (Spoken Language Understanding) encounters the challenge of multiple intents, presenting a unique and worthwhile area of research. \cite{mullick2023intent, mullick2022fine, mullick2023novel, mullickexploring, mullick2022evaluation} explore intent detection in different directions. 
AGIF~\cite{qin2020agif}, GL-GIN~\cite{qin2021gl}, \cite{gangadharaiah2019joint}, \cite{song2022enhancing} work on multiple intent identification problem but these approaches do not detect the sentence spans related to different intents and also do not distinguish the primary and non-primary intents. Based on Convert~\cite{henderson2019convert} backed framework, \cite{coope2020span} extract spans for different slots but does not extract and identify multiple intents. \cite{mullick2024matscire,guha2021matscie,mullick2022using} focus on entity extraction in different forms. Previous research also includes both pipeline-based approaches \cite{jiang2023spm} and end-to-end methods \cite{ma2021intention, cui2019user, ma2022unitranser}. However, our work is different from the fact that we identify multiple intent spans along with their corresponding fine and coarse labels.

Our work differs from the fact that, we extract multiple intent spans from a given sentence and detect its coarse and fine intent labels. In this paper, we seek to address the following research questions in the field of multi-label multi-class intent detection with span extraction:

\noindent \textbf{1.} 
We introduce a novel multi-label multi-class intent detection dataset (MLMCID-dataset) utilizing a diverse set of existing datasets with various intent sizes in multilingual settings (English and non-English languages), including coarse and fine-grained intent labeling along with primary and non-primary intent marking.

\noindent \textbf{2.} We thereafter, build a pointer network based encoder-decoder framework to extract multiple intent spans from the given query.


\noindent \textbf{3.} We propose a feed-forward network based intent detection module (MLMCID - \textbf{M}ulti-\textbf{L}abel \textbf{M}ulti-\textbf{C}lass \textbf{I}ntent \textbf{D}etection) to automatically detect multiple primary and non-primary intents for coarse and fine categories in a sextuplet form. We evaluate the performance of MLMCID for full and few shot-settings across several MLMCID datasets. 

\noindent \textbf{4.} We experiment with different LLMs (Llama2, GPT) to assess their efficacy, comparing them with our approach, and providing a detailed qualitative analysis along with a specialized loss function for multi-label multi-class intent detection. 

Empirical findings on various MLMCID datasets demonstrate that our pointer network based RoBERTa model surpasses other baselines methods including LLMs, achieving a higher accuracy with an improvement in macro-F1. 

\section{Dataset}

We conduct different experiments to evaluate our framework on various datasets - all of which are benchmark datasets in NLU domain. We consider three different sizes of the datasets (as per intent class count - mentioned within bracket) - 

(i) \textit{Small}: a) SNIPS (10 intents) \cite{coucke2018snips}, b) ATIS (21 intents) \cite{tur2010left}, c) Facebook Multi-lingual (12 intents) \cite{schuster2018cross} (consisting of the comparable corpus of English, Spanish and Thai data), abbreviated as Fb. 

(ii) \textit{Medium}: a) HWU (64 intents) \cite{liu2019benchmarking}, b) BANKING (77 intents) \cite{casanueva2020efficient}. 

(iii) \textit{Large}: a) CLINC (150 intents) \cite{larson2019evaluation}. 

 Intents of similar domains which convey a similar broader meaning and  are manually grouped together to make coarse-grained labels from original fine-grained labels
\footnote{Course intent is a combination of multiple similar meaning or closely matching finer intents of higher hierarchy. One coarse-grained intent is a cluster of multiple closely matching fine-grained labels.}. 
Table \ref{tab:course-fine-data-fb} shows an example of Facebook-English (Fb-en) combining multiple fine intents (like - `cancel reminder', `set reminder', `show reminders') which are closely similar and convey similar broader meaning of `reminder\_service’ so these are grouped together to form one single broad coarse grained intent label -
`reminder\_service’ and an example of SNIPS combining multiple fine intents (like - `GetTrafficInformation', `ShareETA') are merged into one single course intent class (`Traffic\_update'). Finally, we end up with course intent class of 4 for SNIPS, 5 for Facebook, 18 for HWU, 12 for Banking and 120 for CLINC\footnote{For ATIS we keep fine intents as it is, without coarse intents due to high dis-similarity among intents}. Due to space shortage, the details are in Appendix Table \ref{appendix_snips_banking_annotation} and \ref{appendix_fb_clinc_hwu_annotation}.

\begin{table} [th]
\small
\begin{center}
\begin{tabular}{|p{0.25\textwidth}|p{0.15\textwidth}|}
\hline
  \textbf{Fine Intents Combined} & \textbf{Coarse Intent} \\
 \hline
   cancel reminder, set reminder, show reminders & reminder\_service \\ 
 \hline
 GetTrafficInformation, ShareETA & Traffic\_update\\
 \hline
 \end{tabular}
 \vspace{-1mm}
\caption{Fine-Course Intent for Fb-en and SNIPS}
\label{tab:course-fine-data-fb}
\end{center}
\vspace{-2mm}
\end{table}

All the above datasets are of single intent. In order to validate the broad applicability of the model, we follow the MixAtis and MixSNIPS data-generation guidelines \cite{qin2020agif} to prepare multi-intent datasets for Fb, HWU, BANKING and CLINC. We also use MixATIS and MixSNIPS datasets~\cite{qin2020agif}. All datasets are in English except for Facebook - which contains Spanish and Thai also along with English. 
Three annotators are selected after several discussions and conditions of fulfilling criteria like annotators should have domain
knowledge expertise along with a good working
proficiency in English. Each formed sentence instance is manually checked for correctness, coherence, grammatically meaningful and filter out many sentences which do not qualify. 
Annotators mark Multiple intents and their respective spans within the specified sentence. Annotators also point out which intent is \textit{Primary}\footnote{Between two intents, we define one as primary which is more important than others and main focus of the sentence} and which one is \textit{non-Primary}. If \textit{Primary} and \textit{non-Primary} intents can not be distinguished then both of the intents are considered as \textit{Primary}. 

\begin{table}[!ht]
    \centering
    \begin{adjustbox}{width=0.75\linewidth}
    \begin{tabular}{|c|c|c|c|}
        \hline
        \textbf{Dataset} & \textbf{Train} & \textbf{Dev} & \textbf{Test}\\\hline
Mix-SNIPS &	11000 & 2197 &	2198\\\hline
Mix-ATIS&	13161 &	600 &	829\\\hline
FB-EN	&800 &	100 &	100\\\hline
FB-ES	&800 &100	&100\\\hline
FB-TH	&800 &	100 &	100\\\hline
HWU64	&780	&97	&97\\\hline
BANKING	&1156&	144&	144\\\hline
CLINC	&1353	&169	&169\\\hline
Yahoo &	498 &	62 &	162 \\\hline
MPQA & 284	& 36 &	136  \\\hline
    \end{tabular}
    \end{adjustbox}
    \vspace{-1mm}
    \caption{\textbf{MLMCID-dataset} statistics}
    \label{tab:fine-intent-class}
    \vspace{-3mm}
\end{table}

To show the real world applicability of our framework,  we also experiment on two different practical datasets: a) MPQA\footnote{https://mpqa.cs.pitt.edu/} (Multi Perspective Question Answering)~\cite{mullick2016graphical,mullick2017generic}, b) Yahoo News article \cite{mullick2016graphical,mullick2017generic}. 
Intent can be broadly categorised as opinionated or factual. Each sentence from MPQA and Yahoo news articles is marked as opinion and fact. Further, opinions can be of four different subcategory \cite{asher2009appraisal} - `Report', `Judgment', `Advise' and `Sentiment' and facts can be subcategorised into five types \cite{soni2014modeling} - `Report', `Knowledge', `Belief', `Doubt' and `Perception'. So coarse intent can be sub-categorized in four opinionated fine-intents and five factual fine-intents. In MPQA and Yahoo news article, annotators are told to identify different clauses of compound and complex sentences and mark the fine label intent categories for opinion and fact. In all the annotation tasks - initial labeling is done by
two annotators and any annotation discrepancy is
checked and resolved by the third annotator after discussing with others. Overall inter-annotator agreement is 0.89 which is considered good as per \cite{landis1977measurement}. The detail statistics of train-dev-test divisions of different dataset intent dataset are shown in Table \ref{tab:fine-intent-class}. We term this dataset as \textbf{MLMCID-dataset.}

\begin{figure*}[!t]
    \centering
    \includegraphics[width=\textwidth]{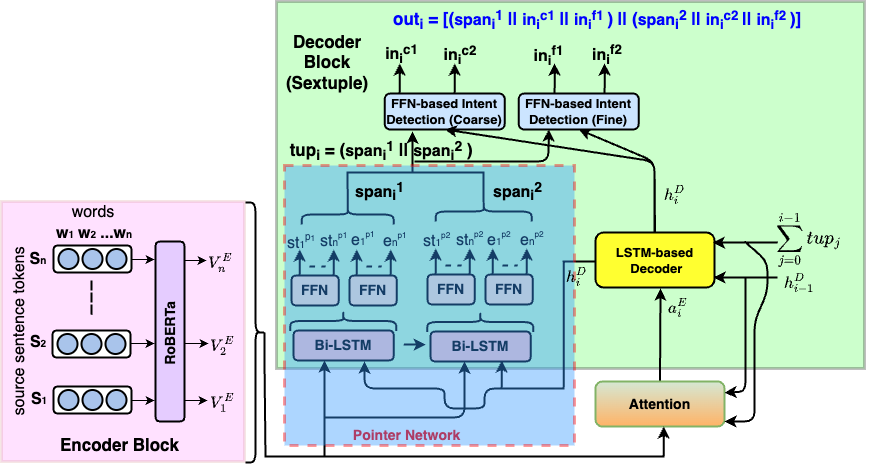}
    \caption{Pointer Network Based multi-label, multi-class intent detection (MLMCID) architecture}
    \label{fig:PMLITE_model}
    \vspace{-2mm}
\end{figure*}

We use the Facebook data from \textbf{MLMCID-dataset} comprising 1000 text instances and corresponding intent labels are annotated for its 3 variations - English, Spanish and Thai. The text instances of English, Spanish and Thai languages are termed as Facebook (English), Facebook (Spanish) and Facebook (Thai) dataset respectively.

\section{Problem Definition}

To formally describe the multi-label, multi-class intent detection (MLMCID) problem setting, let there be an input sentence $S_{i}$ = $\{w_{1}, w_{2}, ..., w_{n}\}$ contains $n$ words. The model aims to extract multiple intent spans along with their coarse and fine classes in the form of a sextuple, $ST = \{out_{i} | out_{i} = [(st_{i}^{p_{1}}, e_{i}^{p_{1}}), in_{i}^{c_{1}}, in_{i}^{f_{1}}, (st_{i}^{p_{2}}, e_{i}^{p_{2}}), in_{i}^{c_{2}}, in_{i}^{f_{2}}]\}_{i=1}^{|ST|}$; 

where $t_{i}$ denotes the $i^{th}$ triplet and $|ST|$ denotes the length of the sextuple set. 
$st_{i}^{p_{1}}$ and $st_{i}^{p_{2}}$ represents the beginning position of first intent span and second intent span respectively for the $i^{th}$ sextuple. Similarly, $e_{i}^{p_{1}}$ and $e_{i}^{p_{2}}$ denotes the end position of first intent span and second intent span for the $i^{th}$ sextuple. So ($st_{i}^{p_{1}}$ and $e_{i}^{p_{1}}$) mark the first intent span for the $i^{th}$ sextuple. Similarly, ($st_{i}^{p_{2}}$ and $e_{i}^{p_{2}}$) mark the second intent span for the $i^{th}$ sextuple. $in_{i}^{c_{1}}$ and $in_{i}^{f_{1}}$ represents the possible coarse and fine intent class of the first intent span. Similarly, $in_{i}^{c_{2}}$ and $in_{i}^{f_{2}}$ represents the possible coarse and fine intent class of the second intent span. $p_1$ and $p_2$ denote the two pointer network models. Pointer Network Model has the following advantages: it is a joint model for entity extraction and relation classification. Pointer network model can detect an intent in a sentence in a form of triplet (intent span, coarse intent label, fine intent label) even if there is an overlap with other intents. $c_1$ and $c_2$ mark the course labels. $f_1$ and $f_2$ indicates fine labels. $out_i$ is the $i^{th}$ output sextuple.

\section{Solution Approach}

For the task of multi-label, multi-class intent detection (MLMCID), our goal is to jointly extract the intent spans along with detecting multiple coarse and fine intents. Our MLMCID output representation is a sextuple format. We employ pointer network based architecture for joint extraction of the sextuple. Following are the different components of solution framework approach:

\subsection{Encoder}

We use four different embeddings in the encoder block (for English language datasets): a) BERT (`bert-base-uncased')~\cite{devlin2019bert}, b) RoBERTa (`roberta-base-uncased')~\cite{liu2019roberta}, c) DistilBERT~\cite{sanh2019distilbert} and d) Electra~\cite{clark2020electra}. For non-English language datasets (Facebook Thai and Spanish), we utilise mBERT (multilingual BERT) \cite{pires2019multilingual}, XLM-R (XLM-RoBERTa) \cite{conneau-etal-2020-unsupervised} and mDistilBERT \cite{sanh2019distilbert}. mBERT architecture pre-trained on Wikipedia articles from 104 languages. 
XLM-RoBERTa is a large multi-lingual language model based on RoBERTa, trained on 2.5TB of filtered CommonCrawl data. mDistilBERT is a distilled version of mBERT containing 134 million parameters.

Let, $S_i$ be the $i^{th}$ sentence containing $w_1$, $w_2$ ... $w_n$ words. After sentence encoding, the encoder generates a vector ($\mathbf{V}_i^E$) from the $i^{th}$ sentence $S_i$. It is shown in the `Encoder Block' in Fig 2.

\subsection{Decoder}

We apply a Pointer Network-based approach along with LSTM-based sequence generator, attention model and FFN (Feed-Forward Network) architecture (Similar to \cite{nayak2020effective}) to identify intent spans and predict the coarse and fine intent labels. Different blocks are as following: 


\noindent \textbf{LSTM-based Sequence Generator:} The sequence generator structure is based on an LSTM layer with hidden dimension $D_h$ to produce the sequence of two intent spans. Using the attention layer sentence encoding ($a_{i}^E$), pointer network based previous tuple ($\mathbf{tup}_{i}$) 
and hidden vectors ($h_{i-1}^D$) as input to generate the hidden representation of the current token ($h_{i}^D$).  
 The $tup_0$ = ($\overrightarrow{0}$) denotes the dummy tuple. Following are LSTM outcomes:


\begin{align}
&\mathbf{tup}_{i}=\sum_{j=0}^{i-1}\mathbf{tup}_{j} \\
&\mathbf{h}_i^D = \mathrm{LSTM}(\mathbf{a}_i^E \Vert \mathbf{tup}_{i-1}, \mathbf{h}_{i-1}^D)
\end{align}

\vspace{-3mm}

\begin{align}
&\hat{st}_i^1 = {w}_{st}^1 {h}_i^m + {b}_{st}^1,
\quad \hat{e}_i^1 = {w}_e^1 {h}_i^m + {b}_e^1\\
&st_{i}^{p_{1}} = \mathrm{softmax}(\hat{st}_i^1),
\quad e_{i}^{p_{1}} = \mathrm{softmax}(\hat{e}_i^1)
\end{align}



\noindent \textbf{Attention Modeling:} Utilizing \citet{bahdanau2014neural} attention algorithm we use previous tuple ($tup_{i-1}$) and hidden vector ($h_{i-1}^D$) as input at timestamp $t$ to produce the attention weighted context vector ($a_{i}^E$) for the current input sentence. 

\noindent \textbf{Pointer Network:} 
A Bi-LSTM layer with hidden dimension $\mathbf{D}_{H}$, followed by two FFN (Feed Forward Networks), constitutes a pointer network. Here we use two-pointer networks for extracting two intent spans. We concatenate $\mathbf{h}_i^D$ and $\mathbf{V}_i^E$ (obtained from the encoding layer) to provide the input of a Bi-LSTM model (forward and backward LSTM), which provides a hidden representation to be fed to FFN models. Two FFNs with softmax provide scores between 0 and 1, the start ($st$) and end ($e$) index of one intent span. 

\noindent where ${w}_{st}^1$ and ${w}_e^1$ are the weight parameters of FFN. ${b}_{st}^1$ and ${b}_e^1$ are the bias parameters of the feed-forward layers (FFN). $\hat{{st}}_i^1$ and $\hat{{e}}_i^1$ are normalized probabilities of the $i^{th}$ source sentence. $st_{i}^{p_{1}}$ and $e_{i}^{p_{1}}$ denotes the begin and end token of the first intent span in the first pointer network model of the $i^{th}$ source sentence. Then, the second pointer network model extracts the second entity. After concatenating the first Bi-LSTM output vector ($\mathbf{h}_i^m$) with decoder sequence generator output ($\mathbf{h}_i^D$) and sentence encoding ($\mathbf{V}_i^E$), we feed them to the second pointer network to obtain the position of the begin and end tokens of the second intent span. Together, these two pointer networks produce the feature vectors $tup_{i}$ containing intent span 1 ($span_i^1$) and span 2 ($span_i^2$).

\noindent \textbf{Intent Detector:} We concatenate $\mathbf{tup}_{i}$ with $\mathbf{h}_i^D$ and pass
it through a feed-forward network (FFN) with softmax to produce the normalized probabilities over
intent sets and thereby predict the coarse ($in_{i}^{c_{1}}$, $in_{i}^{c_{2}}$) and fine ($in_{i}^{f_{1}}$, $in_{i}^{f_{2}}$) intent labels for first and second spans.

\begin{table*} [!ht]
\scriptsize
\begin{center}
\begin{adjustbox}{width=\textwidth}
\begin{tabular}{|c|c|c|c|c|c|c|c|c|}
\hline
  \multicolumn{2}{|c|}{\textbf{Dataset}} & \textbf{BERT (p, av)} & \textbf{RoBERTa (p, av)} & \textbf{DistilBERT (p, av)} & \textbf{Electra (p, av)} & \textbf{Llama2 (p, av)} & \textbf{GPT-3.5 (p, av)} & \textbf{GPT-4 (p, av)} \\
  \hline
  \multirow{2}{0.1\textwidth}{MIX\_SNIPS} & A & 
  89.2,80.2 & \textbf{90.0},\textbf{81.9} & 89.2,80.2 & 89.8,80.7 & 48.3,41.2 & 60.4,55.8 & 64.7,61.1\\
  \cline{2-9}
  & F1 & 89.0,80.1 & \textbf{89.7},\textbf{82.1} & 88.5,79.4 & 89.5,80.5 & 42.6,40.5 & 60.2,56.2 & 62.5,60.3\\
   \hline
   \multirow{2}{0.1\textwidth}{FACEBOOK (English)} & A & 98.0,80.8 & \textbf{98.5},\textbf{81.2} & 97.2,80.2 & 97.4,80.5 & 21.0,19.2 & 70.7,62.1 & 75.6,76.5\\
  \cline{2-9}
  & F1 & \textbf{98.2},\textbf{88.2} & 92.8,82.8 & 92.8,82.2 & 92.8,83.1 & 20.6,19.6 & 65.3,60.8 & 72.6,70.5\\
   \hline
   \multirow{2}{0.1\textwidth}{MIX\_ATIS} & A & 71.3,64.6 & \textbf{70.2},\textbf{63.5} & 72.2,63.6 & 70.6,59.7 & 16.9,15.0 & 29.5,32.5 & 38.7,32.8\\
  \cline{2-9}
  & F1 & 51.7,38.6 & \textbf{53.4},\textbf{38.8}  & 50.3,35.8 & 46.3,35.5 & 15.7,14.0 & 27.2,31.5 & 36.8,32.6\\
  \hline
  \multirow{2}{0.1\textwidth}{HWU64} & A & 83.5,68.0 & \textbf{85.5},\textbf{70.0} & 82.5,66.2 & 83.0,66.2 & 35.8,38.1 & 56.0,52.3 & 59.1,53.1\\
  \cline{2-9}
  & F1 & 81.9,65.9 & \textbf{80.0},\textbf{63.7} & 79.9,64.1 & 79.4,62.5 & 32.9,30.5 & 50.6,51.2 & 57.3,56.4\\
   \hline
   \multirow{2}{0.1\textwidth}{BANKING} & A & 84.0,76.9 & \textbf{85.4},\textbf{78.5} & 78.8,70.9 & 79.9,71.8 & 31.5,31.6 & 25.4,20.5 & 47.9,47.4\\
  \cline{2-9}
  & F1 & 82.7,71.4 & \textbf{85.2},\textbf{75.2} & 79.2,67.9 & 79.4,68.1 & 28.2,29.1 & 20.2,20.3 & 45.2,43.6\\
   \hline
   \multirow{2}{0.1\textwidth}{CLINC} & A & 86.3,72.7 & \textbf{92.3},\textbf{81.3} & 79.8,68.0 & 88.7,71.7 & 57.5,55.9 & 58.7,57.2 & 64.3,56.6\\
  \cline{2-9}
  & F1 & 77.1,64.1 & \textbf{88.3},\textbf{75.5} & 71.7,60.0 & 81.3,63.0 & 51.2,50.3 & 56.3,55.3 & 63.7,54.3\\
   \hline
   \multirow{2}{0.1\textwidth}{Overall Average} & A & 84.1,75.7 & \textbf{88.2,78.5} & 82.2,73.2 & 85.7,72.2 & 34.1,37.0 & 49.2,38.1 & 60.6,53.3\\
  \cline{2-9}
  & F1 & 80.8,73.9 & \textbf{85.2,75.8} & 81.4,70.6 & 80.9,71.3 & 30.5,32.8 & 44.9,41.4 & 58.7,53.6\\
   \hline
\end{tabular}
\end{adjustbox}
\caption{\label{coarse_performance} Overall Accuracy (A) and Macro F1-score (F1) in (\%) of different models in \textbf{MLMCID} and LLMs for coarse labels (on English Datasets) - primary intent (p) and average(av). (The best outcomes are marked in \textbf{Bold})}
\end{center}
\end{table*}

\subsection{Baselines}
We employ different open-source LLMs with prompt based fine-tuning on the training set to generate the two different intent spans and detect coarse and fine intents. 

\noindent \textbf{Llama2:}\footnote{\url{https://ai.meta.com/llama/}} We apply Llama2-7b~(\cite{touvron2023llama}) using Quantized Low-Rank Adaptation (QLoRA) \cite{dettmers2023qlora} (to optimize training efficiency) for supervised fine-tuning using MLMCID-Datasets. 

\noindent \textbf{GPT:} We also use state-of-the-art large-size LLMs, developed by OpenAI: GPT-3.5~\cite{gpt35turbo}~\footnote{https://chat.openai.com/} and GPT-4~\cite{gpt4}\footnote{https://openai.com/gpt-4} with example based prompting to extract intent spans and identify coarse and fine intents (Computed on April, 2024).

\begin{table*} [!b]
\scriptsize
\begin{center}
\begin{adjustbox}{width=\textwidth}
\begin{tabular}{|c|c|c|c|c|c|c|c|c|}
\hline
  \multicolumn{2}{|c|}{\textbf{Dataset}} & \textbf{BERT (p, av)} & \textbf{RoBERTa (p, av)} & \textbf{DistilBERT (p, av)} & \textbf{Electra (p, av)} & \textbf{Llama2 (p, av)} & \textbf{GPT-3.5 (p, av)} & \textbf{GPT-4 (p, av)} \\
  \hline
  \multirow{2}{0.1\textwidth}{MIX\_SNIPS} & A & 85.4,80.9 & \textbf{89.6},\textbf{85.0} & 87.5,81.9 & 86.3,80.9 & 35.0,20.1 & 64.2,60.5 & 64.7,61.1\\
  \cline{2-9}
  & F1 & 83.5,80.1 & \textbf{89.0},\textbf{85.9} & 86.6,81.7 & 86.2,82.1 & 27.5,22.1 & 55.6,51.2 & 57.3,54.9 \\
   \hline
   \multirow{2}{0.1\textwidth}{FACEBOOK (English)} & A & 96.5,81.3 & 97.5,80.7 & 96.5,79.7 & \textbf{98.5},\textbf{81.7} & 11.1,12.1 & 44.4,46.4 & 73.4,77.6\\
  \cline{2-9}
  & F1 & 87.5,79.5 & 94.5,82.0 & 78.4,73.1 & \textbf{95.4},\textbf{82.7} & 9.2,9.7 & 40.2,41.3 & 69.5,69.8 \\
   \hline
   \multirow{2}{0.1\textwidth}{MIX\_ATIS} & A & 71.3,64.6 & \textbf{70.2},\textbf{63.5} & 72.2,63.6 & 70.6,59.7 & 16.9,15.0 & 29.5,32.5 & 38.7,32.8\\
  \cline{2-9}
  & F1 & 51.7,38.6 & \textbf{53.4},\textbf{38.8}  & 50.3,35.8 & 46.3,35.5 & 15.7,14.0 & 27.2,31.5 & 36.8,32.6\\
  \hline
   \multirow{2}{0.1\textwidth}{HWU64} & A & 74.1,57.2 & \textbf{83.0},\textbf{67.1} & 75.1,57.7 & 70.1,53.9 & 29.8,20.3 & 41.8,33.2 & 52.5,48.2\\
  \cline{2-9}
  & F1 & 57.9,43.6 & \textbf{68.3},\textbf{52.8} & 61.0,44.6 & 54.5,41.6 & 25.6,19.6 & 31.6,30.5 & 48.9,46.3\\
   \hline
  \multirow{2}{0.1\textwidth}{BANKING} & A & 78.5,61.2 & \textbf{82.3},\textbf{71.2} & 69.5,54.3 & 73.3,57.2 & 19.0,17.7 & 21.0,20.5 & 27.3,25.7 \\
  \cline{2-9}
  & F1 & 73.5,57.0 & \textbf{80.0},\textbf{68.4} & 64.1,51.4 & 67.8,52.4 & 15.6,16.2 & 18.1,19.4 & 25.6,24.3\\
   \hline
   \multirow{2}{0.1\textwidth}{CLINC} & A & 88.1,73.9 & \textbf{89.3},\textbf{81.2} & 81.6,68.1 & 84.9,70.8 & 43.0,37.8 & 47.0,40.9 & 55.7,48.1\\
  \cline{2-9}
  & F1 & 81.7,66.9 & \textbf{85.3},\textbf{74.2} & 75.2,60.8 & 79.4,63.4 & 39.6,35.7 & 45.4,39.5 & 51.2,45.3\\
   \hline
   \multirow{2}{0.1\textwidth}{Overall Average} & A & 82.3,69.9 & \textbf{85.3},\textbf{74.8} & 80.4,67.5 & 80.6,67.4 & 25.8,20.5 & 41.3,39.0 & 52.1,48.9 \\
   \cline{2-9}
   & F1 & 72.7,60.9 & \textbf{78.4},\textbf{66.9} & 69.3,57.9 & 71.6,59.7 & 22.2,19.6 & 36.4,35.6 & 48.2,45.5\\
   \hline
\end{tabular}
\end{adjustbox}
\caption{Overall Accuracy (A) and Macro F1-score (F1) in (\%) of different models in \textbf{MLMCID} and LLMs for fine labels (on English Datasets) - primary intent (p) and average(av). (The best outcomes are marked in \textbf{Bold})}
\label{fine_performance}
\end{center}
\end{table*}

\begin{table*} [!ht]
\scriptsize
\begin{center}
\begin{adjustbox}{width=\textwidth}
\begin{tabular}{|c|c|c|c|c|c|c|c|c|}
\hline
  \multicolumn{3}{|l|}{\textbf{Dataset}} & \textbf{mBERT (p, av)} & \textbf{XLM-R (p, av)} & \textbf{mDistilBERT (p, av)} & \textbf{Llama-2 (p, av)} & \textbf{GPT-3.5 (p, av)} & \textbf{GPT-4 (p, av)} \\
  \hline
  \multirow{4}{0.08\textwidth}{FACEBOOK (Spanish)} & \multirow{2}{0.05\textwidth}{Coarse} & A & 98.0,80.7 & \textbf{98.5},\textbf{81.5} & 98.0,80.2 & 51.2,39.9 & 64.6,61.6 & 70.7,75.6\\
  \cline{3-9}
  & & F1 & 91.3,82.2 & \textbf{92.5},\textbf{82.7} & 91.1,82.9 & 47.2,39.6 & 62.6,61.3 & 69.4,69.3 \\
   \cline{2-9}
  & \multirow{2}{0.05\textwidth}{Fine} & A & 96.7,80.0 & \textbf{97.5},\textbf{81.0} & 96.5,80.2 & 38.3,27.2 & 57.6,56.6 & 69.7,74.2\\
  \cline{3-9}
  & & F1 & 84.6,80.0 & \textbf{86.0},\textbf{81.7} & 84.3,76.8 & 36.2,30.6 & 55.4,55.0 & 66.2,65.6\\
   \hline
   \multirow{4}{0.08\textwidth}{FACEBOOK (Thai)} & \multirow{2}{0.05\textwidth}{Coarse} & A & 96.5,79.8 & \textbf{97.0},\textbf{80.0} & 96.8,79.0 & 28.0,24.2 & 69.7,58.6 & 73.4,71.5\\
  \cline{3-9}
  & & F1 & 88.4,75.8 & \textbf{96.6},\textbf{78.8} & 94.2,73.4 & 25.6,24.8 & 67.8,57.2 & 71.6,69.3\\
   \cline{2-9}
    &  \multirow{2}{0.05\textwidth}{Fine} & A & 96.0,79.5 & \textbf{96.5},\textbf{79.7} & 95.5,77.2 & 16.3,15.2 & 18.2,18.7 & 68.7,64.9\\
  \cline{3-9}
  & & F1 & 84.1,74.2 & \textbf{82.5},\textbf{75.5} & 68.8,62.7 & 15.7,14.9 & 17.9,16.8 & 59.2,58.7\\
  \hline
  \multirow{4}{0.08\textwidth}{Average} & \multirow{2}{0.05\textwidth}{Coarse} & A & 97.2,80.3 & \textbf{97.8},\textbf{80.8} & 97.4,79.6 & 39.6,32.1 & 67.2,60.1 & 72.1,73.6\\
  \cline{3-9}
  & & F1 & 89.8,79.0 & \textbf{94.6},\textbf{80.8} & 92.6,78.2 & 36.4,32.2 & 65.2,59.3 & 70.5,69.3\\
   \cline{2-9}
    &  \multirow{2}{0.05\textwidth}{Fine} & A & 97.3,79.7 & \textbf{97.0},\textbf{80.8} & 96.0,78.7 & 27.3,21.2 & 37.9,37.7 & 69.2,69.6\\
  \cline{3-9}
  & & F1 & 84.3,77.1 & \textbf{84.3},\textbf{78.6} & 76.5,69.8 & 25.9,22.8 & 36.7,35.9 & 62.7,62.2 \\
   \hline
\end{tabular}
\end{adjustbox}
\caption{Overall Accuracy (A) and Macro F1 (F1) in (\%) of different models in \textbf{MLMCID} and LLMs for coarse and fine grained labels of Facebook Spanish and Thai datasets - primary intent (p) and overall average(av). (The best outcomes are marked in \textbf{Bold})}
\label{multilingual_fb_performance}
\end{center}
\vspace{-3mm}
\end{table*}

\section{Experiments}


To validate our proposed framework, we compare the Pointer Network Model (PNM) of MLMCID while taking various embeddings as input: BERT, RoBERTa, DistilBERT, and Electra  on all datasets. We also explore different large language models (Llama2-7b, GPT-3.5 and GPT-4) to check how effectively they can extract multiple intent spans and detect different intents. After that, we experiment with different variations of overall best performing RoBERTa model - varying the training data size to understand how much training data is required for decent performance. We also perform zero-shot and few-shot experiments to check the approach's usefulness in the presence of minimal data. Tables \ref{coarse_performance}, \ref{fine_performance} and \ref{multilingual_fb_performance} show the overall performances of different models for the English (Mix-SNIPS, Mix-ATIS, Facebook, HWU, BANKING and CLINC) and Non-English (Facebook Thai and Spanish) datasets. We use prediction accuracy and macro F1-score as evaluation metrics. Table \ref{coarse_performance} and \ref{fine_performance} infer performances on primary and overall average of coarse and fine intent labels on English datasets. Following are the details of our findings:

\noindent \textbf{Findings 1:} For coarse label intent detection, as shown in Table \ref{coarse_performance}, RoBERTa (with PNM) in MLMCID achieves superior performances in terms of accuracy and F1-score across all datasets of different intent sizes (Mix-SNIPS, Mix-ATIS, HWU, BANKING, CLINC) for both primary intent detection and overall average except for Facebook English where BERT is more effective in terms of F1-score for both primary and overall average. 

\noindent \textbf{Findings 2:} Similar to coarse intent detection, for fine label intent detection, RoBERTa (with PNM) in MLMCID also produce better results than others in terms of accuracy and F1-score for most of the cases across all English datasets except for Facebook English dataset, where Electra provides better outcome in terms of accuracy and F1-score for both primary and overall intent detection. It is shown in Table \ref{fine_performance}. 

\noindent \textbf{Findings 3:} For all English datasets, BERT, RoBERTa, DistilBERT and Electra performs almost similar with decent accuracy and F1-score which signifies the utility of pointer network model based \textbf{MLMCID}  architecture. 

\begin{table*} [!b]
\scriptsize
\begin{center}
\begin{adjustbox}{width=0.9\textwidth}
\begin{tabular}{|c|c|c|c|c|c|c|c|c|}
\hline
    \multirow{2}{3pt}{\textbf{Th}} & \multicolumn{8}{|c|}{\textbf{Dataset (primary (p) and average (av) intent) in \%}} \\
    \cline{2-9}
    & \textbf{MIX\_SNIPS} & \textbf{FB\_en}  & \textbf{FB\_es} & \textbf{FB\_th} & \textbf{MIX\_ATIS} & \textbf{HWU64} & \textbf{BANKING} & \textbf{CLINC}  \\
    \hline
    50 \% & 89.2,80.9 & 96.0,78.5 & 94.5,77.4 & 89.9,82.4 & 95.1,90.2 & 85.5,70.0 & 81.8,74.7 & 90.1,79.2\\
    \hline
    60 \% & 87.7,78.9 & 95.0,77.9 & 86.5,71.2 & 77.4,70.3 & 91.9,90.2 & 85.5,68.9 & 79.4,72.0 & 88.4,77.5 \\
    \hline
    70 \% & 79.4,70.8 & 91.0,74.6 & 75.6,63.1 & 75.2,67.7 & 85.1,89.2 & 84.6,68.1 & 75.9,68.3 & 84.0,73.0\\
    \hline
    80 \% & 70.4,63.5 & 83.0,68.8 & 72.6,59.4 & 71.4,62.9 & 83.8,88.2 & 81.9,66.6 & 69.9,62.8 & 79.1,67.6 \\
    \hline
    90 \% & 59.2,54.2 & 75.0,63.2 & 61.6,50.3 & 69.4,59.6 & 80.9,86.2 & 77.5,62.6 & 63.4,56.0 & 67.5,58.2 \\
    \hline
\end{tabular}
\end{adjustbox}
\caption{\label{coarse_performance_pmlite} Overall Accuracy (A) in (\%) of RoBERTa model in \textbf{MLMCID} for coarse grained labels (on English Datasets) - primary (p) and average (av) intents. (`Th' indicates threshold value) }
\end{center}
\end{table*}

\begin{table*} [!b]
\scriptsize
\begin{center}
\begin{adjustbox}{width=0.9\textwidth}
\begin{tabular}{|c|c|c|c|c|c|c|c|c|}
\hline
    \multirow{2}{3pt}{\textbf{Th}} & \multicolumn{8}{|c|}{\textbf{Dataset (primary (p) and average (av) intent) in \%}} \\
    \cline{2-9}
    & \textbf{MIX\_SNIPS} & \textbf{FB\_en}  & \textbf{FB\_es} & \textbf{FB\_th} & \textbf{MIX\_ATIS} & \textbf{HWU64} & \textbf{BANKING} & \textbf{CLINC} \\
    \hline
    50 \% & 83.6,80.7 & 93.5,78.1 & 91.5,75.9 & 89.6,81.1 & 95.1,90.2 & 83.0,67.1 & 77.1,69.8 & 86.6,78.9\\
    \hline
    60 \% & 82.1,78.9 & 92.5,77.0 & 85.6,70.2 &  82.4,79.6 & 91.9,90.2 & 80.4,65.0 & 74.8,67.5 & 86.1,77.4\\
    \hline
    70 \% & 76.1,72.3 & 87.6,71.9 & 78.7,63.8 & 75.9,67.2 & 85.1,89.2 & 79.5,64.3 & 69.1,62.2 & 82.9,70.9\\
    \hline
    80 \% & 68.6,64.8 & 78.7,65.9 & 74.8,60.6 & 68.4,61.0 & 83.8,88.2 & 75.2,62.4 & 64.5,56.0 & 77.0,68.0\\
    \hline
    90 \% & 55.2,52.4 & 72.8,61.0 & 63.0,50.7 & 65.4,57.3 & 80.9,86.2 & 67.5,55.1 & 57.7,49.4 & 66.4,62.8\\
    \hline
\end{tabular}
\end{adjustbox}
\caption{\label{fine_performance_pmlite} Overall Accuracy (A) in (\%) of RoBERTa model in \textbf{MLMCID} for fine grained labels (on English Datasets) - primary (p) and average (av) intents. (`Th' indicates threshold value) }, 
\end{center}
\end{table*}

\begin{table*} [!ht]
\scriptsize
\begin{center}
\begin{adjustbox}{width=\textwidth}    
\begin{tabular}{|p{0.06\textwidth}|p{0.05\textwidth}|p{0.12\textwidth}|p{0.10\textwidth}|p{0.10\textwidth}|p{0.10\textwidth}|p{0.10\textwidth}|p{0.12\textwidth}|p{0.10\textwidth}|}
\hline
  \multicolumn{2}{|c|}{\textbf{Dataset}} & \textbf{Llama2-7b Fine-tune (p,av)} & \textbf{Llama2-7b Vanilla (p, av)} & \textbf{GPT-3.5 (p, av)} & \textbf{GPT-4 (p, av)} & \textbf{RoBERTa-SNIPS(p, av)} & \textbf{RoBERTa-BANKING(p,av)} & \textbf{RoBERTa-CLINC(p, av)}\\
  \hline
  \multirow{2}{0.1\textwidth}{MPQA} & {Fine} & 42.8,27.1 & 18.8,16.9 & 20.0,14.2 & 48.5,37.1 & 45.0,42.5 & 44.5,42.0 & 43.9,41.5\\
  \cline{2-9}
    & {Coarse} & 65.7,64.2 & 51.9,50.0 & 62.8,59.9 & 68.5,45.6 & 75.6,43.7 & 73.0,41.9 & 72.8,42.6 \\
   \hline
    \multirow{2}{0.1\textwidth}{YAHOO} & {Fine} & 48.3,37.5 & 18.8,15.8 & 11.4,10.6 & 58.0,56.2 & 55.3,54.9 & 54.0,53.8 & 52.9,54.2\\
  \cline{2-9}
    & {Coarse} & 61.2,49.9 & 52.8,50.0 & 50.0,50.0 & 61.2,49.1 & 66.3,65.7 & 64.5,62.9 & 63.2,60.8\\
   \hline
\end{tabular}
\end{adjustbox}
\caption{\label{llm_performance} Overall Accuracy (A) in (\%) of RoBERTa model in \textbf{MLMCID} (trained on SNIPS, BANKING and CLINC) and LLMs for fine and course grained labels - primary (p) and average (av) intent.  }
\end{center}
\end{table*}

\noindent \textbf{Findings 4:} We observe that the LLMs (Llama-2-7b, GPT-3.5, GPT-4) fall behind in performance from Pointer Network based approaches with different encoders, even though they are much larger than our proposed framework, thus strengthening the need for such a specialized \textbf{MLMCID} architecture. Llama2-7b performs poorly among three LLMs - this may be due to the fact of less contextual understanding in this specific task. More details in Appendix \ref{appendix:experimental_findings}.

\noindent \textbf{Findings 5:} RoBERTa with PNM in MLMCID performs better than any other models for overall average accuracy and F1-score across all English datasets for both primary and average course and fine intent detection after intent spans extraction. 


\noindent \textbf{Findings 6:} For non-English languages like Spanish (Facebook) and Thai (Facebook) datasets , we observe that for both fine and coarse grained intent labels, XLM-R and mBERT both produce good results but XLM-R outperforms mBERT in all aspects across all datasets and overall for both primary intent detection and overall average intent detection with intent span extraction.


\noindent \textbf{Findings 7:} To check the effectivity of span extraction by pointer network, we vary the similarity (extracted intent span vs actual intent span) threshold utilise that extracted span to check the overall accuracy. We check for 50\% - 90\% similarity threshold range and overall framework (RoBERTA with PNM) accuracies (for both primary and average intent) across all datasets for coarse and fine intent labels are shown in Table \ref{coarse_performance_pmlite} and \ref{fine_performance_pmlite}. It is seen a good performance even with 50\% similarity which shows the efficacy of the system.

\subsection*{Ablation Studies}

\noindent \textbf{1. K-shot setting:} To evaluate the RoBERTa based PNM model of MLMCID architecture, we utilize K samples for all English datasets where K = 5 (5-shot) and 10 (10-shot) for coarse and fine intent labels. The accuracy and F1-score of primary and average intents are shown in Table \ref{tab:k-shot}. This shows even with very limited number of data-points (like in 5-shot), the system is able to achieve a decent performance across different datasets.

\noindent \textbf{2. Practical Datasets:} We test the trained RoBERTa models with PNM (using SNIPS, BANKING and CLINC dataset) in MLMCID to evaluate on external MPQA and Yahoo datasets. We also check LLMs - Llama2-7b (vanilla and finetuned), GPT-3.5 and GPT-4 on MPQA and Yahoo but RoBERTa based PNM in MLMCID outperfomrs LLMs in most of the cases and show decent performance as shown in Table \ref{llm_performance}. It is seen that, for Llama2-7b vanilla model performs poorly and fine-tune version perform better but does not outperform GPT and RoBERTa based models.

\noindent \textbf{3. Intent Counts:} All datasets have two intents (primary and non-primary) in one sentence except for Yahoo, 2.6\% cases with more than 2 intents so we show all results considering the case of 2 intents in a sentence. Our system is also effective for more than two intents by utilizing more pointer network block in the decoder framework, as shown in Appendix \ref{appendix:intent_greater_than_two}.

\begin{table} [!ht]
\scriptsize
\begin{adjustbox}{width=\linewidth}
\begin{tabular}
{|c|c|c|c|c|}
\hline
  \multicolumn{3}{|l|}{\textbf{Dataset}} & \textbf{Coarse (p, avg)} & \textbf{Fine (p, avg)} \\
  \hline
  \multirow{4}{0.1\textwidth}{SNIPS} & \multirow{2}{0.05\textwidth}{5-shot} & A & 61.0,49.2 &70.9,53.3 \\ 
  \cline{3-5}
  & & F1 & 58.1,46.4 & 67.9,51.7\\
  \cline{2-5}
  & \multirow{2}{0.05\textwidth}{10-shot} & A & 61.4,52.1 & 75.9,63.1\\
  \cline{3-5}
  & & F1 & 60.7,47.4 & 75.1,61.0\\ 
   \hline
   \multirow{4}{0.1\textwidth}{FACEBOOK (English)} & \multirow{2}{0.05\textwidth}{5-shot} & A & 83.5,62.0 & 76.0,58.3\\ 
  \cline{3-5}
  & & F1 & 58.0,42.8 & 26.7,20.4\\
  \cline{2-5}
  & \multirow{2}{0.05\textwidth}{10-shot} & A & 87.5,67.8 & 83.5,64.3\\
  \cline{3-5}
  & & F1 & 59.5,45.9 & 34.3,25.2\\ 
   \hline
   \multirow{4}{0.1\textwidth}{HWU-64} & \multirow{2}{0.05\textwidth}{5-shot} & A & 57.2,39.3 & 47.8,29.6\\ 
  \cline{3-5}
  & & F1 & 49.3,34.7 & 35.5,22.1\\
  \cline{2-5}
  & \multirow{2}{0.05\textwidth}{10-shot} & A & 62.2,43.5 & 62.2,43.3\\
  \cline{3-5}
  & & F1 & 58.2,39.2 & 46.2,31.9\\ 
   \hline
   \multirow{4}{0.1\textwidth}{BANKING} & \multirow{2}{0.05\textwidth}{5-shot} & A & 36.0,28.2 & 62.3,38.6\\ 
  \cline{3-5}
  & & F1 & 32.5,25.0 & 56.7,34.4\\
  \cline{2-5}
  & \multirow{2}{0.05\textwidth}{10-shot} & A & 46.0,32.9 & 76.1,52.9\\
  \cline{3-5}
  & & F1 & 46.1,31.4 & 71.2,48.0\\ 
  \hline
  \multirow{4}{0.1\textwidth}{CLINC} & \multirow{2}{0.05\textwidth}{5-shot} & A & 78.4,50.4 & 76.3,53.4\\ 
  \cline{3-5}
  & & F1 & 69.9,44.0 & 65.8,44.6\\
  \cline{2-5}
  & \multirow{2}{0.05\textwidth}{10-shot} & A & 87.3,65.9 & 89.6,69.7\\
  \cline{3-5}
  & & F1 & 79.3,58.5 & 79.3,58.5\\ 
   \hline
\end{tabular}
\end{adjustbox}
\caption{Accuracy (A) and F1-Score for coarse and fine\\ intents by RoBERTa(in \%) for k-shot, k = \{5, 10\}  }
\label{tab:k-shot}
\end{table}

\noindent \textbf{Experimental Settings:} Our experiments are conducted on two Tesla P100 GPUs with 16 GB RAM, 6 Gbps clock cycle, GDDR5 memory and one 80GB A100 GPU, 210MHz clock cycle, 2*960 GB SSD with 5 epochs. We use Adam optimizer with learning rate: $10^{-5}$ with cross-entropy as the loss function, weight decay: $10^{-5}$ and a dropout rate of 0.5 is applied on the embeddings to avoid overfitting for all experiments (Details are in Appendix). All methods took less than 120 GPU minutes 
(except Llama2: $\sim$4-5 hrs) for fine tuning and $\sim$2 hrs for inference. All the hyperparameters are tuned on the dev set. We have used NLTK, Spacy, Scikit-learn, openai (version=0.28), huggingface\_hub, torch and transformers python packages for all experiments and evaluation \footnote{All Code / Data details are in \url{https://github.com/ankan2/multi-intent-pointer-network}}.


\section{Loss Function}

We calculate loss of different intent classes across all samples for primary, non-primary intents and their respective primary and non primary spans as shown in equation \ref{eqn:primary-loss}, \ref{eqn:non-primary-loss} and \ref{eqn:span-loss} respectively. 
For training our model, we minimize the sum of negative log-likelihood loss for classifying the intent and the four pointer locations corresponding to the primary and non primary intent spans as shown in equation \ref{eqn:span_loss_multiline}.

    
{\small
\begin{equation}
    \mathcal{L}_{p} = - \frac{1}{N}\sum_{i=1}^{N} \Big{[}\sum_{j=1}^{C} (y_{1})_{ij}log(p_{ij})- \frac{1}{J} \sum_{j=1}^{J} \log ((y_{1})j^n)\Big{]}
\label{eqn:primary-loss}
\end{equation}
}
{\small
\begin{equation}
    \mathcal{L}_{np} = - \frac{1}{N}\sum_{i=1}^{N} \Big{[}\sum_{j=1}^{C} (y_{2})_{ij}log(p_{ij})- \frac{1}{J} \sum_{j=1}^{J} \log ((y_{2})j^n)\Big{]}
\label{eqn:non-primary-loss}
\end{equation}
}

\begin{equation}
\begin{split}
    \mathcal{L}_{span} = - \frac{1}{N \times J} \sum_{n=1}^{N} \sum_{j=1}^{J} \Big{[}&\log ((st^{p_{1}}){j}^n \cdot \\(e^{p{1}}){j}^n)
        + \log ((st^{p{2}}){j}^n \cdot (e^{p{2}}){j}^n)\Big{]}
\end{split}
\label{eqn:span-loss}
\end{equation}

\begin{figure}[t]
     \centering
     \begin{subfigure}[b]{0.475\linewidth}
         \centering
         \includegraphics[width=0.95\linewidth]{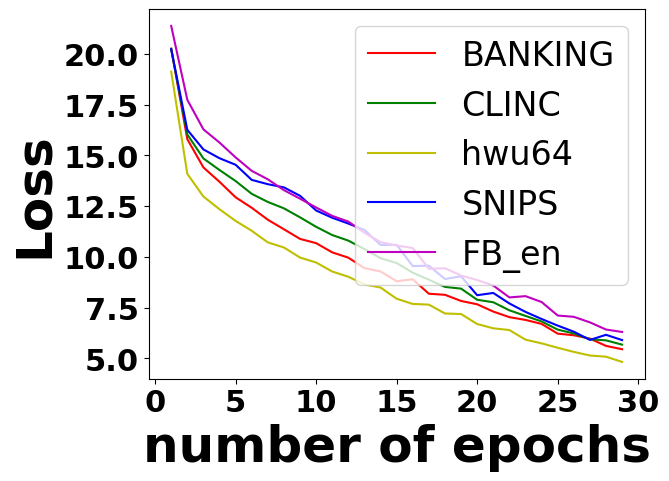}
         \caption{Combined loss - Coarse}
         \label{coarse_loss}
     \end{subfigure}
     \begin{subfigure}[b]{0.475\linewidth}
         \centering
         \includegraphics[width=0.95\linewidth]{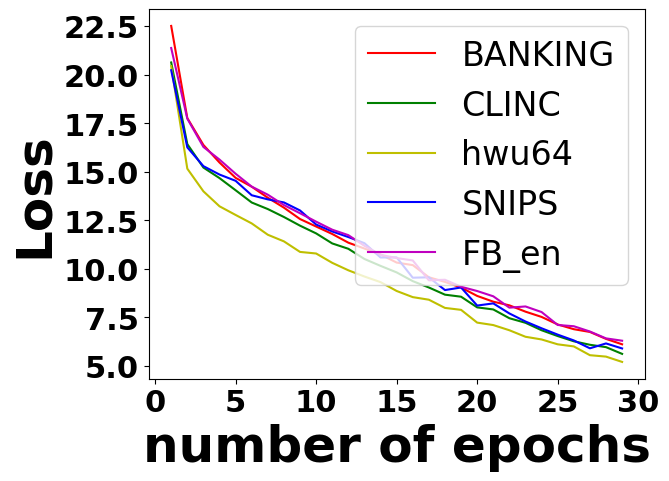}
         \caption{Combined Loss - Fine}
         \label{fig:fine_loss}
     \end{subfigure}
     \caption{By RoBERTa based pointer network \\ (PNM) model in \mlmcid}
     \vspace{-4mm}
     \label{fig:loss_performance}
\end{figure}

\noindent Here, $C$ is the number of intent classes and $(y_{1}) \in \{in^{c1}, in^{f1}\}$ and $(y_{2}) \in \{in^{c2},in^{f2}\}$. $(y_{1}){ij}$ and $(y_{2}){ij}$ are the one-hot ground truth labels for sample $i$ and class $j$ for the primary and non-primary intents respectively, and $p_{ij}$ is the predicted probability for sample $i$ and class $j$. $n$ represents the $n^{th}$ training instance with $N$ being the batch size, $j$ represents the $j^{th}$ decoding time step with $J$ being the length of the longest target sequence among all instances in the current batch. $st^p, e^p; \;p \in \{p_{1}, p_{2}\}$ respectively represent the softmax scores corresponding to the true start and end positions of the primary and non primary spans. Fig \ref{fig:loss_performance} shows the variation of the overall loss for course and fine intents with respect to the training progress (in terms of epochs) across different datasets. Loss decreases with larger epochs and after 10 epochs the loss  decrement is significant to obtain decent outcome.

\begin{equation}
    \begin{split}
        \mathcal{L} = \mathcal{L}_p + \mathcal{L}_{np} + \mathcal{L}_{span}
    \end{split}
    \label{eqn:span_loss_multiline}
\end{equation}

\section{Conclusion}


Intent detection is crucial in task-oriented conversation systems. Earlier works focus on scenarios with the presence of a single intent and do not extract intent spans. This work is one of the first to consider multiple intents in a single sentence within a conversation system, including primary and non-primary intents. First, we create novel datasets using state-of-the-art datasets with coarse and fine intent labels. Then, we develop a Pointer Network-based encoder-decoder framework (MLMCID - multi-label multi-class intent detection) using RoBERTa (for English data) and XLM-R (for non-English data) to jointly extract intent spans from sentences and detect corresponding coarse and fine intents. We show that the MLMCID model even outperforms various LLMs for these specific tasks across different datasets. The approach demonstrates efficacy even in few-shot scenarios. Qualitative analysis shows a reasonable grasp of primary and secondary intent concepts. Overall, this highlights the importance of multi-intent modeling for real-world conversational AI, with the datasets and models providing a strong foundation for future research.

\section*{Limitations and Discussion}
Table \ref{coarse_performance}, \ref{fine_performance}, \ref{multilingual_fb_performance} shows that even when our model fails to give the correct predictions exactly, it predicts the primary intent correctly most of the time. This is due to the fact we are using the top-2 intents to infer the primary and non-primary intents using the same classifier. 
Also, in some examples, the primary and non-primary intent Labels, when predicted wrongly, are swapped, suggesting that the model is still able to grasp the notion of intent. We shall work on these limitations in future. 

\section*{Ethical Concerns}

We use publicly available codes and datasets so there is no ethical concerns.

\section*{Acknowledgements}
The work was supported in part by Prime Minister Research Fellowship (PMRF).

\bibliography{anthology,custom}
\bibliographystyle{acl_natbib}

\pdfoutput=1








%
%


\title{Instructions for ACL 2023 Proceedings}


\author{First Author \\
  Affiliation / Address line 1 \\
  Affiliation / Address line 2 \\
  Affiliation / Address line 3 \\
  \texttt{email@domain} \\\And
  Second Author \\
  Affiliation / Address line 1 \\
  Affiliation / Address line 2 \\
  Affiliation / Address line 3 \\
  \texttt{email@domain} \\}


\appendix

\newpage
\newpage

\section{Experimental Findings} \label{appendix:experimental_findings}

\subsection{Why encoder decoder model performs well} 
\label{appendix:findings_8}

Pointer Network model is a state-of-the-art approach which is ideal for extracting multiple spans from a sentence using the pointing mechanism to directly select positions in the input sequence, allowing for variable-length outputs and precise boundary identification. Their attention mechanism effectively handles context, enabling accurate span extraction in a computationally efficient manner. It is effective also because of - 
\begin{itemize}
    \item Dynamically predict entity spans within a sequence, enhancing adaptability across various NLP tasks
    \item capture the interdependence between spans and intents, crucial for tasks where one intent's prediction relies on another characteristics within the same context. 
    \item Reduce the need for manual feature engineering, learning to predict spans directly from input data for more efficient models
    \item Finally, enable end-to-end learning by directly predicting entity span positions, facilitating seamless integration with other neural network components.
\end{itemize}

\subsection{PNM for more than two intent cases} 
\label{appendix:intent_greater_than_two}


To evaluate the effectiveness of the Pointer Network framework for more than two intents, we experimented with a small sample from the MIX\_SNiPS, BANKING, and CLINC datasets, incorporating three intents. For instance, the sentence "Will it snow this weekend? Please help me book a rental car for Nashville and play that song called 'Bring the Noise'" includes the intents: {weather, car\_rental, play\_music}. Table \ref{tab:3-intents-study} presents the performance of RoBERTa on this annotated sample. The results demonstrate the effectiveness of our system in handling a larger number of intents, as reflected by the accuracy (in \%).

\begin{table}[]
\begin{adjustbox}{width=\columnwidth}
\begin{tabular}{|c|c|c|c|c|}
\hline
Dataset & Intent 1 (\%)	& Intent 2 (\%) &	Intent 3 (\%) &	Average (\%)   \\ \hline
MIX\_SNIPS (fine) &	81.2 &	73.8 & 60.3 &	71.7 \\ \hline
MIX\_SNIPS (coarse) &	85.4 &	74.4 &	62.3 &	74.0 \\ \hline
BANKING (fine) &	79.3 &	60.0 &	56.3 &	65.2 \\ \hline
BANKING (coarse) &	83.3 &	68.9 &	59.6 &	70.6 \\ \hline
CLINC (fine) &	80.7 &	69.2 &	55.4 &	68.4 \\ \hline
CLINC (coarse) &	81.9 &	71.7 &	58.3 &	70.6 \\ \hline
\end{tabular}
\end{adjustbox}
\caption{3-Intent Detection by Roberta based PNM}
\label{tab:3-intents-study}
\end{table}


\subsection{Scalability} \label{appendix:scalability}

We experiment with datasets composed of two intents with the P100 server with 16GB GPU \ref{appendix:exp_settings} where 6-9 GB GPU VRAM has been utilised. Further we experiment on the dataset with three intents in the same server which use 12-13 GB GPU VRAM so our approach is scalable and applicable in resource constrained environments. It is also seen that in case of larger numbers of intents with the introduction of additional pointer networks - the system is scalable and does not require large computational costs. So the framework can be useful in real time processing for large scale systems. Though it is also to be noted that most of the datasets are composed with two intents even in the real life sentences.

\subsection{Single Intent Detection} \label{single_intent}

We perform additional experiments on three  datasets with various intent sizes - SNIPS (small), BANKING (medium) and CLINC (large) and detect the single-intent text using RoBERTa based pointer network architecture - which is shown in the following table (in \%). It shows the effectiveness of our model for coarse (c) and fine (f).

\begin{table}[!ht]
\centering
\begin{adjustbox}{width=0.7\columnwidth}
\begin{tabular}{|c|c|c|}
\hline
Dataset &	coarse (\%) &	fine (\%) \\\hline
SNIPS &	90.0 &	85.9  \\\hline
BANKING &	83.9 &	81.8  \\\hline
CLINC &	80.0 &	75.3  \\\hline
\end{tabular}
\end{adjustbox}
\caption{Single Intent Detection}
\label{tab:single-intent-study}
\end{table}

\section{Experimental Settings} \label{appendix:exp_settings}

Our experiments are conducted on two Tesla P100 GPUs with 16 GB RAM, 6 Gbps clock cycle, GDDR5 memory and one 80GB A100 GPU, 210MHz clock cycle, 2*960 GB SSD with 5 epochs. We use Adam optimizer with learning rate: $10^{-5}$ with cross-entropy as the loss function, weight decay: $10^{-5}$ and a dropout rate of 0.5 is applied on the embeddings to avoid overfitting for all experiments. All methods took less than 120 GPU minutes (except Llama2: $\sim$4-5 hrs) for fine tuning and $\sim$2 hrs for inference. All the hyperparameters are tuned on the dev set. We have used NLTK, Spacy, Scikit-learn, openai(version=0.28), huggingface\_hub, torch and transformers python packages for all experiments and evaluation.

\section{Example} \label{appendix:example}

\begin{figure*}[th]
    \centering
    \includegraphics[width=1\textwidth]{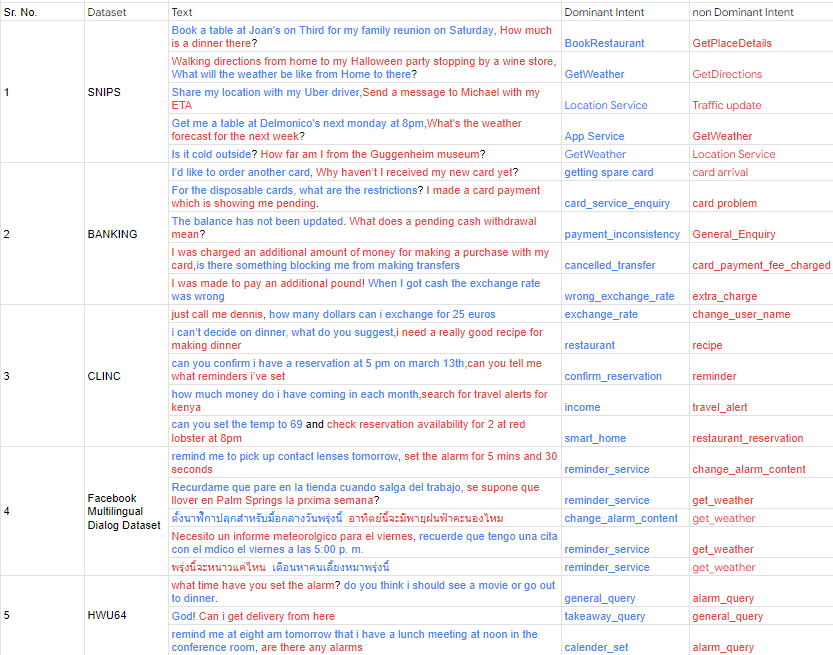}
    \caption{Examples in \textcolor{black}{\mlmcid} Dataset}
    \label{appendix_dataset_snapshot}
\end{figure*}

Figure \ref{appendix_dataset_snapshot} shows some examples from \textbf{MLMCID} dataset. Table \ref{appendix_snips_banking_annotation} and \ref{appendix_fb_clinc_hwu_annotation} shows some examples of fine to coarse label conversion for \textbf{MLMCID} dataset. Table \ref{appendix_wrong_pred} shows some examples of the intent classes predicted with their respective confidence for PNM.

\begin{table*} [th]
\small
\begin{center}
\begin{tabular}{|p{0.02\textwidth}|p{0.12\textwidth}|p{0.2\textwidth}|p{0.4\textwidth}|}
\hline
  \textbf{Sr. No.} & \textbf{Dataset} & \textbf{Coarse Label} & \textbf{Fine Labels Combined} \\
 \hline
 \multirow{2}{0.02\textwidth}{1.} & \multirow{2}{*}{SNIPS} & Traffic\_update & ComparePlaces, GetPlaceDetails, ShareCurrentLocation, SearchPlace, GetDirections\\ 
 \cline{3-4}
 & & App\_Service & RequestRide, BookRestaurant\\ 
 \cline{3-4}
 & & Location\_service & GetTrafficInformation, ShareETA\\
 \cline{3-4}
 & & GetWeather & GetWeather\\
 \hline
 \multirow{12}{0.02\textwidth}{2.} & \multirow{12}{*}{BANKING} & Cancelled\_ transfer & cancel\_transfer, beneficiary\_not\_allowed\\ 
 \cline{3-4}
 & & Card\_problem & card\_arrival, card\_linking, card\_swallowed, activate\_my\_card, declined\_card\_payment, reverted\_card\_payment?, pending\_card\_payment, card\_not\_working, lost\_or\_stolen\_card, pin\_blocked, card\_payment\_fee\_charged, card\_payment\_not\_recognised, card\_acceptance\\ 
 \cline{3-4}
 & & exchange\_rate\_query & exchange\_rate, fiat\_currency\_support, card\_payment\_wrong\_exchange\_rate, wrong\_exchange\_rate\_for\_cash\_withdrawal\\ 
 \cline{3-4}
 & & General\_Enquiry & extra\_charge\_on\_statement, card\_delivery\_estimate, pending\_cash\_withdrawal, automatic\_top\_up, verify\_top\_up, topping\_up\_by\_card, exchange\_via\_app, atm\_support, lost\_or\_stolen\_phone, transfer\_timing, transfer\_fee\_charged, receiving\_money, top\_up\_by\_cash\_or\_cheque, exchange\_charge, cash\_withdrawal\_charge, apple\_pay\_or\_google\_pay\\ 
 \cline{3-4}
 & & Top\_up & top\_up\_by\_bank\_transfer\_charge, pending\_top\_up, top\_up\_limits, top\_up\_reverted, top\_up\_failed\\ 
 \cline{3-4}
 & & Account\_opening & age\_limit\\ 
 \cline{3-4}
 & & transaction\_problem & contactless\_not\_working, wrong\_amount\_of\_cash\_received, transfer\_not\_received\_by\_recipient, balance\_not\_updated\_after\_cheque\_or\_cash\_deposit, declined\_cash\_withdrawal, pending\_transfer, transaction\_charged\_twice, declined\_transfer, failed\_transfer\\ 
 \cline{3-4}
 & & Card\_service\_enquiry & visa\_or\_mastercard, disposable\_card\_limits, getting\_virtual\_card, supported\_cards\_and\_currencies, getting\_spare\_card, virtual\_card\_not\_working, top\_up\_by\_card\_charge, card\_about\_to\_expire, country\_support\\ 
 \cline{3-4}
 & & Identity\_verification & unable\_to\_verify\_identity, why\_verify\_identity, verify\_my\_identity\\ 
 \cline{3-4}
 & & Service\_request & order\_physical\_card, edit\_personal\_details, get\_physical\_card, passcode\_forgotten, change\_pin, terminate\_account, request\_refund, verify\_source\_of\_funds, transfer\_into\_account, get\_disposable\_virtual\_card\\ 
 \cline{3-4}
 & & Malpractice & compromised\_card, cash\_withdrawal\_not\_ recognised\\ 
 \cline{3-4}
  & & Payment\_inconsistency & direct\_debit\_payment\_not\_recognised,  Refund\_not\_showing\_up,  balance\_not\_updated\_after\_bank\_transfer\\ 
  \hline
  \end{tabular}
\caption{\label{appendix_snips_banking_annotation}Fine to Coarse Labels Conversion Examples for SNIPS and BANKING Dataset}
\end{center}
\end{table*}
  
\begin{table*} [th]
\small
\begin{center}
\begin{tabular}{|p{0.02\textwidth}|p{0.12\textwidth}|p{0.2\textwidth}|p{0.4\textwidth}|}
\hline
  \textbf{Sr. No.} & \textbf{Dataset} & \textbf{Coarse Label} & \textbf{Fine Labels Combined} \\
 \hline
  \multirow{6}{0.02\textwidth}{3.} & \multirow{6}{*}{CLINC} & health\_suggestion & nutrition\_info, oil\_change\_how, calories\\ 
 \cline{3-4}
 & & Restaurant & restaurant\_reviews, accept\_reservations, restaurant\_reservation, meal\_suggestion, restaurant\_suggestion\\ 
 \cline{3-4}
 & & account & redeem\_rewards, report\_lost\_card, balance, bill\_balance, credit\_limit, rewards\_balance, bill\_due, credit\_score, transactions, spending\_history, damaged\_card, pin\_change, replacement\_card\_duration, new\_card, direct\_deposit, credit\_limit\_change, payday, application\_status, pto\_request, pto\_request\_status, pto\_balance, pto\_used\\ 
 \cline{3-4}
 & & communication & make\_call, text\\ 
 \cline{3-4}
 & & Reminder & remind\_update, remind, reminder\_update, reminder, meeting\_schedule\\ 
 \cline{3-4}
 & & banking\_enquiry & account\_blocked, freeze\_account, interest\_rate\\ 
 \hline
  \multirow{5}{0.02\textwidth}{4.} & \multirow{5}{0.12\textwidth}{Facebook Multilingual Dialog Dataset} &  change\_alarm\_content & cancel alarm, modify alarm, set alarm, snooze alarm\\ 
  \cline{3-4}
 & & reminder\_service & cancel reminder, set reminder, show reminders\\ 
 \cline{3-4}
 & & sunset\_sunrise & weather check sunrise, weather check sunset\\ 
 \cline{3-4}
 & & get\_weather & weather find\\
  \cline{3-4}
  & & read alarm content & show alarm, time left on alarm\\
  \hline
  \multirow{18}{0.02\textwidth}{5.} & \multirow{18}{0.12\textwidth}{HWU64} & alarm & set, remove, query\\ 
  \cline{3-4}
 & & audio & audio\_volume\_mute, audio\_volume\_down, audio\_volume\_other, audio\_volume\_up\\ 
 \cline{3-4}
 & & iot & iot\_hue\_lightchange, iot\_hue\_lightoff, iot\_hue\_lighton, iot\_hue\_lightdim, iot\_cleaning, iot\_hue\_lightup, iot\_coffee, iot\_wemo\_on, iot\_wemo\_off\\ 
 \cline{3-4}
 & & calendar & calendar\_query, calendar\_set, calendar\_remove\\
  \cline{3-4}
  & & play & play\_music, play\_radio, play\_audiobook, play\_podcasts, play\_game\\
  \cline{3-4}
 & & general & general\_query, general\_greet, general\_joke, general\_negate, general\_dontcare, general\_repeat, general\_affirm, general\_commandstop, general\_confirm, general\_explain, general\_praise\\ 
 \cline{3-4}
 & & datetime & datetime\_query, datetime\_convert\\ 
 \cline{3-4}
 & & takeaway & takeaway\_query, takeaway\_order\\
  \cline{3-4}
  & & news & news\_query\\
  \cline{3-4}
 & & music & music\_likeness, music\_query, music\_settings, music\_dislikeness\\ 
 \cline{3-4}
 & & weather & weather\_query\\ 
 \cline{3-4}
 & & qa & qa\_stock, qa\_factoid, qa\_definition, qa\_maths, qa\_currency\\
  \cline{3-4}
  & & social & social\_post, social\_query\\
  \cline{3-4}
  & & recommendation & recommendation\_locations, recommendation\_events, recommendation\_movies\\
  \cline{3-4}
 & & cooking & cooking\_recipe, cooking\_query\\ 
 \cline{3-4}
 & & email & email\_sendemail, email\_query, email\_querycontact, email\_addcontact\\ 
 \cline{3-4}
 & & transport & transport\_query, transport\_ticket, transport\_traffic, transport\_taxi\\
  \cline{3-4}
  & & lists & lists\_query, lists\_remove, lists\_createoradd\\
  \hline
 \end{tabular}
\caption{\label{appendix_fb_clinc_hwu_annotation}Fine to Coarse Labels Conversion Examples for Facebook and CLINC Dataset}
\end{center}
\end{table*}

\begin{table*}[th]
\small
\begin{center}
\begin{tabular}{|p{0.23\textwidth}|p{0.24\textwidth}|p{0.2\textwidth}|p{0.165\textwidth}|p{0.065\textwidth}|} 
 \hline
  \textbf{Text} & \textbf{Predicted} & \textbf{True Label} & \textbf{Remarks about prediction}\\ 
 \hline
 Find a store near Sia's place where I can buy champagne and find me a brunch spot in Lower Manhattan (SNIPS) & Location\_Service (Primary), App\_Service (Non-Primary) & Location\_Service, Location\_Service & Non-Primary Label predicted wrongly\\
 \hline
 Book a cab, is there traffic on the US 50 portion I'm going to take to go to my client meeting? (SNIPS) & App\_Service
 (Primary), Traffic\_update (Non-Primary) & Traffic\_update, App\_Service & Wrong Predictions - swapped ground-truth labels\\
 \hline 
 What will the weather be like at my Airbnb this week end? Is there a parking at my hotel? (SNIPS) & GetWeather (Primary),	Location\_Service (Non-Primary) & GetWeather, Location\_Service & Correct Predictions\\
 \hline 
 Can you make a reservation at a lebanese restaurant nearby, for lunch, party of 5? How's the traffic from here?
 (SNIPS) & App\_Service (Primary), Traffic\_update (Non-Primary) & App\_Service, Location\_Service & Non-Primary label wrongly predicted\\
 \hline 
 set alarm,remind me to pay electric monday (FACEBOOK) & set alarm (Primary), set reminder (Non-Primary) & set alarm, set reminder & Correct Predictions\\
 \hline
 is it going to snow in chicago tomorrow, any chance of rain today? (FACEBOOK) & weather find (Primary), set reminder (Non-Primary) & weather find, weather find & Non-Primary label wrongly predicted\\
 \hline
 how hot will it be, how long will it rain tomorrow (FACEBOOK) & weather find (Primary), set reminder (Non-Primary) & weather find, weather find & Non-Primary label wrongly predicted\\
 \hline
 what is the average wait for transfers, I'm still waiting on my identity verification.(BANKING)	& General\_Enquiry (Primary), Identity\_verification (Non-Primary) & General\_Enquiry, Identity\_verification & Correct Predictions\\
 \hline
 My card is due to expire,Why can't I get cash out (BANKING) & card\_about\_to\_expire (Primary), declined\_cash\_withdrawal (Non-Primary) & card\_about\_to\_expire, declined\_cash\_withdrawal & Correct Predictions\\
 \hline
 I have a new email. I am in the EU. Can I get one of your cards? (BANKING) & Card\_service\_enquiry (Primary), General\_Enquiry (Non-Primary) & Service\_request, Card\_service\_enquiry & Incorrect Predictions; Predicted Primary Intent is same as the Non-Primary Ground Truth Label\\
 \hline
 Can other people top up my account? where did my funds come from? (BANKING) & verify\_source\_of\_funds (Primary), topping\_up\_by\_card (Non-Primary) & topping\_up\_by\_card, verify\_source\_of\_funds & Wrong Predictions - swapped ground-truth labels\\
 \hline
 Can you tell me my shopping list items, please? Is tomato on my shopping list? (CLINC) & shopping\_list (Primary), account (Non-Primary) & shopping\_list, shopping\_list & Non-Primary label wrongly predicted\\
 \hline
 Change the name of your system. Your name from this point forward is george. (CLINC) & change\_ai\_name (Primary), change\_user\_name (Non-Primary) & change\_ai\_name, change\_ai\_name & Non-Primary label wrongly predicted\\
 \hline
 use my phone and connect please,tell me something that'll make me laugh(CLINC) & sync\_device (Primary), tell\_joke (Non-Primary) & sync\_device, tell\_joke & Correct Predictions\\
 \hline
 will there be traffic on the way to walmart,can you help me with a rental car(CLINC) & traffic (Primary), car\_rental (Non-Primary) & traffic, car\_rental & Correct Predictions \\
 \hline
\end{tabular}
\caption{Prediction of best-performing models and Respective Confidence}
\label{appendix_wrong_pred}
\end{center}
\end{table*}

\end{document}